\documentclass[a4paper]{article}

\usepackage[english]{babel}
\usepackage[utf8]{inputenc}
\usepackage{amsmath}
\usepackage{graphicx}
\usepackage[colorinlistoftodos]{todonotes}

\title{Latest Datasets and Technologies Presented in the Workshop on Grasping and Manipulation Datasets}

\author{Matteo Bianchi, Jeannette Bohg, and Yu Sun
\thanks{Authors contributed equally to this manuscript and are listed in alphabetic order.}
}
\begin{document}
\maketitle

\begin{abstract}
This paper reports the activities and outcomes in the Workshop on Grasping and Manipulation Datasets that was organized under the International Conference on Robotics and Automation (ICRA) 2016. The half day workshop was packed with nine invited talks, 12 interactive presentations, and one panel discussion with ten panelists. This paper summarizes all the talks and presentations and recaps what has been discussed in the panels session. This summary servers as a review of recent developments in data collection in grasping and manipulation. Many of the presentations describe ongoing efforts or explorations that could be achieved and fully available in a year or two. The panel discussion not only commented on the current approaches, but also indicates new directions and focuses. The workshop clearly displayed the importance of quality datasets in robotics and robotic grasping and manipulation field. Hopefully the workshop could motivate larger efforts to create big datasets that are comparable with big datasets in other communities such as computer vision. 

\end{abstract}

\section{Introduction}
The Workshop on Grasping and Manipulation Datasets was proposed in September of 2015 and then selected and announced among 14 other workshops in January 2016. It is a half-day workshop that focused on promoting open-access grasping and manipulation related datasets and identified the critical needs on new datasets and a methodology on utilizing the data. 

The workshop is an answer to the recent worldwide trend on providing and using high-quality open-access grasping and manipulation datasets. Many different datasets were recently collected by a number of groups and shared for different research purposes. The datasets are in several different but related research domains with various points of interests and modalities. They include human motion datasets, instrumental activities of daily living (IADL) datasets, other activity datasets, object geometry and motion datasets, haptic interaction datasets, as well as other datasets on human and robot grasping and manipulation. The datasets are not only crucial for evaluating and comparing the performances of novel methods, but also extremely valuable for offline robotic learning and training.  

The objective of the workshop is to address the need to promote the use of open-access datasets, coordinating the efforts and resources and avoid pitfalls in collecting high-quality datasets, clearing up confusion in selecting suitable datasets, and identifying new demands in datasets. The workshop brought together researchers from different domains for their common interests in grasping and manipulation related datasets. 

The workshop was comprised of two oral sessions, one interactive session, and panel discussion. The presentations in the oral sessions are invited and the presentations in the interactive session are openly solicited and reviewed to attract broader participants and facilitate vibrant discussions. The oral session presentations were concise and limited to 15 minutes. The interactive presentations included a two-min oral presentation and a 45-minute poster presentation. In the panel session, open questions were discussed that were formulated ahead of time or raised during the talks. 

This article summarizes what was reported in the workshop and provides an updated overview of the latest landscape of grasping and manipulation datasets.

\section{Invited presentations}

The workshop consisted of nine invited presentations on different aspects of building and using datasets for studying human and robotic grasp. Various sensing modalities and analysis tools were considered.
The list of speakers was:
\begin{itemize}
\item Aaron Dollar (Yale University): The YCB object benchmark for manipulation research;
\item Yasemin Bekiroglu (University of Birmingham): Assessing grasp stability and object shape modeling based on visual and tactile data;
\item Matteo Bianchi (University of Pisa and Istituto Italiano di Tecnologia):  An open-access repository to share data and tools for the study of human and robotic hands: the HandCorpus initiative
\item Jeannette Bohg (Max Planck Institute): Leveraging Big Data for Grasp Planning
\item Yu Sun (University of South Florida): Interactive motion and wrench in instrument manipulation
\item Tamim Asfour (Karlsruher Institut fur Technologie): The KIT Whole-Body Human Motion Database
\item Hamal Marino and Marco Gabiccini (University of Pisa): Datasets (and tools) from disconnected markers to organized behaviors: a path towards autonomous manipulation
\item Jeffrey Mahler and Ken Goldberg (University of Berkeley): Dexterity Network (Dex-Net): A Cloud-Based Network of 3D Objects for Robust Grasp Planning
\end{itemize}

To summarize the content of talks, we can identify three main topics:
\begin{itemize}
\item[(1)] \textbf{Datasets for robot planning, robotic manipulation research, benchmarking and grasp stability assessment}
\item[(2)] \textbf{Datasets related to human body actions}
\item[(2)] \textbf{Datasets of instruments in interactions with the environment}
\end{itemize}

\subsection{Benchmarking}
Regarding point (1), \textbf{Aaron Dollar} presented \textbf{the Yale-CMU-Berkeley (YCB) Object and Model set} \cite{calli2015benchmarking}, which is intended to be used to facilitate benchmarking in robotic manipulation research. The objects in the set are designed to cover a wide range of aspects of the manipulation problem. The set includes objects of daily life with different shapes, sizes, textures, weight and rigidity, as well as some widely used manipulation tests. The associated database provides high-resolution RGBD scans, physical properties, and geometric models of the objects for easy incorporation into manipulation and planning software platforms. In addition to describing the objects and models in the set along with how they were chosen and derived, a framework and a number of example task protocols are also provided, laying out how the set can be used to quantitatively evaluate a range of manipulation approaches including planning, learning, mechanical design, control, and many others. The set can be distributed to research groups worldwide. The goal of this initiative is to enable the community of manipulation researchers to more easily compare approaches as well as continually evolve standardized benchmarking tests and metrics as the field matures. 

As mentioned in Dollar's work, benchmarking is also crucial to enable a correct usage of RGBD scans and videos. Indeed, the development of such  new sensing technologies capable of providing high quality synchronized videos of both color and depth  can represent a key factor for enhancing robotic object recognition, manipulation, navigation, and interaction capabilities, also due to the high potential for mass adoption. In his talk, \textbf{Dieter Fox} presented the effort in developing the first \textbf{large-scale, hierarchical multi-view RGBD based dataset for object recognition} \cite{lai2011large}. More specifically, the datasets contains 300 objects organized into 51 categories and has been made publicly available to the research community so as to enable rapid progress based on this promising technology. In addition to introducing a large object dataset, Fox also described \textbf{techniques for RGBD based object recognition and detection}, demonstrating that combining color and depth information can substantially improve the results, which can be evaluated at different levels (e.g. Category and Instance level), also targeting classification of  previously unseen objects and with potential implications for grasping and manipulation datasets.

Real sensory data can be used not only to understand object shape, which is important for grasp planning, but also to learn models \textbf{to assess grasp success (discriminative and generative)}. To tackle both these issues, in her talk, \textbf{Yasemine Bekiroglu} described probabilistic approaches with real sensory data, e.g., visual and tactile \cite{bekiroglu2012learning}.
Indeed, an important ability of a robot that interacts with the environment and manipulates objects, is to deal with the uncertainty in sensory data, which is necessary to perform on-line assessment of grasp stability. Bekiroglu presented a method for assessing grasp stability based on haptic data and machine learning methods for different sensory streams, which can  affects the success of grasping process both in the planning stage (before a grasp is executed) and during the execution of the grasp (closed-loop on-line control). Results indicated
that the idea of exploiting the learning approach can be applicable in realistic scenarios. Furthermore, experimental outcomes were presented, which demonstrated that considering both visual and tactile input for stability assessment is beneficial.  Finally, Bekiroglu introduced a \textbf{low-cost pipeline and database for reproducible manipulation research}. This approach combines an inexpensive generation of detailed 3D object models via monocular camera images with a state of the art object tracking algorithm.

\subsection{Grasp Planning}
Back to robotic grasp planning, \textbf{Jeannette Bohg} described a new \textbf{large-scale database of grasps applied to a large set of objects from numerous categories}. Such a database has been publicly released and contains grasps that are generated in simulation and annotated with the standard epsilon and a new physics-metric. Bohg presented a descriptive and efficient representation of the local object shape at which the grasp is applied. Each grasp is annotated with the proposed metrics and representation.

Given these data, a two-fold analysis was considered:
\begin{itemize}
\item \textbf{Crowdsourcing} for the analysis of the correlation of the two metrics with grasp success as predicted by humans. The results confirm that the proposed physics metric is a more consistent predictor for grasp success than the epsilon metric. Furthermore it supports the hypothesis that human labels are not required for good ground truth grasp data. Instead the physics metric can be used for simulation data. 
\item \textbf{Big data learning techniques} (Convolutional Neural Networks and Random Forests) to show how they can leverage the large-scale database for improved prediction of grasp success.
\end{itemize}

Another data driven approach to robot planning and manipulation was presented by \textbf{Jeffrey Mahler} and \textbf{Ken Goldberg}. In this talk, Mahler and Goldberg introduced \textbf{the Dexterity Network (DexNet) 1.0}, a dataset of 3D object models and a sampling-based planning algorithm to explore how Cloud Robotics can be used for robust grasp planning. The algorithm uses a MultiArmed Bandit model with correlated rewards to leverage prior grasps and 3D object models in a growing dataset that currently includes over 10,000 unique 3D object models and 2.5 million parallel-jaw grasps. Each grasp includes an estimate of the probability of force closure under uncertainty in object and gripper pose and friction.  Dex-Net 1.0 uses Multi-View Convolutional Neural Networks (MV-CNNs), a new deep learning method for 3D object classification, to provide a similarity metric between objects, and the Google Cloud Platform to simultaneously run up to 1,500 virtual cores, reducing experiment runtime by up to three orders of magnitude. Experiments suggest that correlated bandit techniques can use a cloud-based network of object models to significantly reduce the number of samples required for robust grasp planning. Code and updated information is available (see \cite{mahlerdex}).

\textbf{Hamal Marino} and \textbf{Marco Gabiccini} discussed how datasets and tools, from disconnected markers to organized behaviors, pushing the attention on \textbf{autonomous robot manipulation}, with special focus on \textbf{soft manipulation}. \textbf{Marino} introduced the envisoned scenario wherethe (not-so-far) future robots will be able to autonomously grasp and manipulate objects, interacting with humans and their environment. Such a scenario has became more and more concrete as the research in the field brings new, promising results, as also discussed in the other talks.

A path started more than 50 years ago, with direct inspection of humans performing various sort of manipulative tasks, passing through categorization (resulting
in the so called “grasp taxonomies”), and then towards building an hardware as close as possible to human hands in order to be able to mimic their behavior. Grasp modeling was used to this end, but the initial hypothesis needed for simplifying such a complex problem has been to have isolated contact points, happening only between the distal phalanxes of the robot hand and an external object. Recently, the concepts of \textbf{soft interaction} and \textbf{soft robotic hands} is becoming increasingly widespread: they started to change the paradigm, from timid, contact-based interactions with the objects to be manipulated, to daring, intense whole–hand interactions also involving the surrounding environment.

Disparate grasp planning algorithms have been developed mainly for the former kind of hands, although some attempts have begun to sprout also for the latter ones.
Moreover, increasing the autonomy of the system, allowing high–level specifications to be interpreted and executed, has been studied as a necessary step towards
simpler, more natural task definition along the path on the way to interaction of the robot with a human–centered world.

In his talk, Marino showed some advancements in aforementioned building blocks, towards the increase of robotic manipulation ability.
The first part dealt with a novel, parametric kinematic model which can be adapted to different subjects and takes into account the relative motion between skin and
bones in order to accurately reconstruct the hand motion of a human performing grasping and manipulation tasks \cite{gabiccini2013data}. Data from various subjects have been collected and analyzed, and a new clustering algorithm is used to obtain a data–driven grasp taxonomy \cite{marino2016datadriven}.

In the second part, the problem of how to transfer the knowledge gathered from humans to robotic systems was faced, and two different ways are explored: recording
humans performing grasping motions while ''wearing'' the robotic end–effector as a tool, and a learning algorithm capable of working with soft robotic hands which
generalizes successful example grasps to new scenarios \cite{marino2016humandriven}. Finally, the third part involved giving the robot an increased autonomy, using an
abstraction layer, which makes robot end–effectors and fixed environment elements alike, each with its own interaction primitives to act on the object; it is shown how it
is possible to translate higher level instructions into a sequence of low level actions, which can be then executed by the robotic system \cite{marino2016problem}.

\subsection{Human activities}
The talk of Hamal Marino is a nice example of \textbf{integration and mutual inspiration between  human studies and robotic applications}. It also underlies the importance of investigating human behavior and the need of benhcmarking protocols for building, using and analyzing datasets of human actions. These aspects were covered by the presentations referring to point (2).

\textbf{Tamim Asfour} presented \textbf{the KIT Whole-Body Human Motion Database} \cite{mandery2015kit}, a publically released large-scale whole-body human motion database consisting of motion data of the observed human subject as well as the objects with which the subject is interacting. Asfour described the procedures for a systematic recording of human motion data with associated complementary data like video recordings and additional sensor measurements (force, IMU, …), as well as environmental elements and objects. The availability of accurate object trajectories together with the associated object mesh models makes the data especially useful for the analysis of manipulation, locomotion and loco-manipulation tasks. Asfour presented procedures and techniques for motion capturing, annotating and organization in large-scale databases as well as for the normalization of human motion to a unified representation based on a reference model of the human body. In addition, he also described methods and software tools for efficient search in the database as well as the transfer of subject-specific motions to robots with different embodiments and discuss several current applications of the database in our current research on whole-body grasping and loco-manipulation

Finally, \textbf{Matteo Bianchi} presented \textbf{HandCorpus}\footnotetext{www.handcorpus.org}, an open-access repository for sharing data, tools and analyses about \textbf{human and robotic hands}. The HandCorpus website represents a cross-platform and user-friendly portal for researchers interested in sharing datasets, analysis tools and/or exchanging ideas, regarding the most versatile end-effector known, the human hand. Over the last years the HandCorpus community has grown and consists now of seven European Committee (EC) projects and more than 20 research groups, located across Europe, Asia and United States of America. Finally, the HandCorpus website is cross-platform, cross-browser and fully accessible through all kind of mobile-devices.

\subsection{Interaction of Instruments}
\textbf{Yu Sun} presented his team's latest work in observing and collecting physical interaction in manipulation tasks. Traditionally, robotic grasping and manipulation approaches have been successful in planning and executing pick-and-place tasks without any physical interaction with other instruments or the environment, which are common in an industry setting. However, when robots work into our daily-living environments and perform a broad range of tasks, all sorts of physical interactions will occur, which will result in physical interactive wrench: force and torque on the instrument in a robot’s hand. The robotic grasp should be able to not only prevent the tools from falling caused by the force of gravity, but also facilitate the interactive wrench \cite{lin2015grasp, lin2016task} and motion \cite{lin2015task, lin2015robot, huang2015generating} that instrument manipulation requires . 

However, compared to the tremendous amount of work done in human motion analysis, very little work has been done in interactive wrench analysis. In the past few years, many manipulation datasets in daily living environments have been collected, but none contains interactive wrench information.

Different from the existing data collection and analysis, Yu Sun's team proposed a physical-interaction observation system that not only observes the motion of the instruments, but also the interactive wrench between the instrument and environment in great detail. The team uses the observation system to collect the instrument motions and wrench measurements in several representative instrument manipulation tasks by a number of participants. The system measures instrument-environment interaction in terms of the 6-DOF interactive wrench (force and torque) between the instrument and the environment using ATI Nano17 and Mini40 F/T sensors and the instruments 6-DOF motion using NaturalPoint OptiTrack MoCap. In addtion, the motion is recorded in RGBD by using a Primesense sensor while the finger motion is recorded by using a 5DT Dataglove. 

The dataset contains 36 manipulations of 25 instruments. Among them, nine are selected from the YBC object sets. Each manipulation takes around 60 seconds and were repeated three times by each of five participants. The dataset is currently only available by request. It will be fully available online once the data collection and processing are finished. 

\section{Interactive Session}
There were 12 interactive presentations. The list was:
\begin{itemize}
\item
Autonomous grasping data collection and tactile signal variability in real-world grasping
Qian Wan, Robert D. Howe

\item
Physically-Consistent Hand Manipulation Dataset
Vikash Kumar, Emo Todorov
\item
Performance Evaluation of 4DoF Gripper Pose Estimation Method by using APC items
Yukiyasu Domae, Ryosuke Kawanishi
\item
Using the YCB Object and Model Set to benchmark the iCub grasping capabilities
Lorenzo Jamone, Alexandre Bernardino, and Jose Santos-Victor
\item
More than a Million Ways to Be Pushed -- A HighFidelity Experimental Data Set of Planar Pushing
Peter K.T. Yu, Alberto Rodriguez, Maria Bauza Villalonga 
\item
Automotive General Assembly Part Datasets And their Environment
Jane Shi
\item
BiGS: BioTac Grasp Stability Dataset
Yevgen Chebotar, Karol Hausman, Zhe Su, Artem Molchanov, Oliver Kroemer, Gaurav Sukhatme, and Stefan Schaal
\item
Recording hand-surface usage in grasp demonstrations
Ravin de Souza, Jose Santos-Victor, and Aude Billard
\item
A dataset of thin-walled deformable objects for manipulation planning
Nicolas Alt, Jingyi Xu and Eckehard Steinbach
\item
CapriDB - Capture, Print, Innovate: A Low-Cost Pipeline and Database for Reproducible Manipulation Research
Florian T. Pokorny*, Yasemin Bekiroglu*, Karl Pauwels, Judith Bütepage, Clara Scherer, and Danica Kragic
\item
Datasets for tactile perception and manipulation
Benjamin Ward-Cherrier, Luke Cramphorn, and Nathan F. Lepora
\item
G3DB: A Database of Successful and Failed Grasps with RGB-D Images, Point Clouds, Mesh Models and Gripper Parameters
Ashley Kleinhans, Benjamin Rosman, Michael Michalik, Bryan Tripp, and Renaud Detry
\end{itemize}

\subsection{Tactile Datasets}
In Qian Wan and Rober D. Howe's presentation ``Autonomous grasping data collection and tactile signal,'' from John A. Paulson School of Engineering and Applied Sciences, Harvard University, a setup was introduced to collect grasping data especially tactile signals autonomously for thousands of grasps. Because the grasping process is in high dimensional space, a large amount of data is usually required for model training testing. The presentation showed an example dataset collected using this setup. The dataset contains raw tactile signals from Takktile Sensors on the robotic fingers of a RightHandRobotics ReFlex Hand in thousands of grasping of a sphere with slightly different sphere locations. 

Benjamin Ward-Cherrier, Luke Cramphorn, and Nathan F. Lepora from the University of Bristol presented another set of tactile interaction databases in their presentation ``Datasets for tactile perception and manipulation.'' The presentation provided an introduction of their unique low cost robust 3d-printed optical tactile sensors called TacTip and TacThumb. The TacTip can achieve 40-fold localization super-resolution to 0.1mm accuracy. The design of both sensors are openly available at opentactip.com. 

In ``Recording hand-surface usage in grasp demonstrations,'' Ravin de Souza, Jose Santos-Victor, and Aude Billard from Ecole Polytechnique Federale de Lausanne (EPFL) and Instituto Superior Tecnico (IST) proposed to cover a data glove with an array of tactile sensors to record tactile interaction between hand surfaces and the grasped objects. The tactile dataglove not only measures 22 joint angles, but also provide contact information through 34 tactile patches. Using the kinematic model of the hand, the contact information on the hand could be reconstructed. They plan to perform a set of studies to obtain tactile information in the hand of grasping all objects in the YCB object set. 

\subsection{Manipulation Datasets}
Peter K.T. Yu, Alberto Rodriguez, and Maria Bauza Villalonga from MIT presented a dataset of friction features in planar pushing in ``More than a Million Ways to Be Pushed -- A High-Fidelity Experimental Data Set of Planar Pushing.'' 
They designed and developed an automatic experiment setup that has an industrial robot performing pushing of various objects along precisely controlled position-velocity-acceleration pushing trajectories. The pushing data are collected in 6
dimensions of variations: surface material, shape of the pushed object, contact position, pushing direction, pushing speed, and pushing acceleration. Dense samples of positions and forces of uniform quality with timestamps are obtained. The obtained data are used to characterize frictional properties at the interaction which are important for accurate computational models. 

In Vikash Kumar and Emo Todorov's presentation ``Physically-Consistent Hand Manipulation Dataset,'' from the University of Washington, a virtual reality (VR) system (MuJoCo Haptix) is presented. It has real-time motion capture, physics simulation, and stereoscopic visualization, so that a user wearing a CyberGlove can manipulate virtual objects in the VR system. Manipulation data including joint kinematics and dynamics, contact interactions, simulated sensor readings could be collected using the physically-realistic simulation in the VR system. The system's low end-to-end latency of 42 msec allows untrained human users to interact with virtual objects in a natural way. The system has been evaluated on a subset of tasks in the Southampton Hand Assessment Procedure (a clinically validated test of hand function). 

\subsection{Datasets for Evaluation}

In ``BiGS: BioTac Grasp Stability Dataset,'' Yevgen Chebotar, Karol Hausman, Zhe Su, Artem Molchanov, Oliver Kroemer, Gaurav Sukhatme, and Stefan Schaal from the University of Southern California presented a dataset of stable and unstable grasping configurations that was collected using the Barrett WAMTM Arm manipulator and the Barrett hand equipped with three BioTac sensors. So far only a cylindrical object is used. The robot reaches down to the object and perform a randomly generated top grasp, and then lifts the object up and performs a range of extensive shaking motions in all directions. If the object is still in the hand after the shaking motions, the grasp is considered to be a stable one. To automate the process, markers are attached to the object and tracked using a VICON motion-tracking system. To bring the object back to up right pose, a fixed bowl is used to catch the object if it is shacked out of the grippers. The setup allows the robot to continuously try and grasp the object for more than 20 hours without human supervision. Total 1000 grasps are collected with 46\% resulted in failures and 54\% succeeded. Raw BioTac electrode values, BioTac pressure sensor values, robot’s joint angles, end-effector pose, object pose obtained from the VICON system, robot’s hand force-torque sensor values, finger strain gauges and finger joint angles are collected in each grasp. The data are used to learn a grasp stability classifier. The details can be found at http://bigs.robotics.usc.edu.

In Yukiyasu Domae and Ryosuke Kawanishi's presentation "Performance Evaluation of 4DoF Gripper Pose Estimation Method by using APC items," from Mitsubishi Electric Corp, the results of grasping the Amazon Picking Challenge (APC) items from bins of a shelf using a two-finger gripper were reported. A set of experiments have been conducted to evaluate their approach that estimates the gripper pose for picking by using their fast graspability evaluation on images from an RGBD sensor. 

Lorenzo Jamone, Alexandre Bernardino, and Jose Santos-Victor from Instituto Superior Tecnico (IST) presented "Using the YCB Object and Model Set to benchmark the iCub grasping capabilities," that reported an evaluation of the iCub capabilities of grasping the YCB Object and Model Set. 
In their study, the iCub fingers are directly controlled by a human expert using a datagloveto to decouple the mechanical capabilities of the
hand from those of a specific controller. With this setup, they was able to determine what objects within the YCB Object and Model Set can be grasped by the iCub and which cannot even with human teleoperation. The results provided a baseline for researchers who want to evaluate the
performance of their grasping controller with the iCub. 

In the presentation ``G3DB: A Database of Successful and Failed Grasps with RGB-D Images, Point Clouds, Mesh Models and Gripper Parameters,'' Ashley Kleinhans, Benjamin Rosman, Michael Michalik, Bryan Tripp, and Renaud Detry from CSIR, the University of Waterloo and the University of Liege presented a database of grasp examples generated in a simulated environment, Virtual Robot Experimentation Platform (V-REP). The group planned to collect grasps of at least a hundred household objects and hundreds of grasps per object in V-REP. A large number of grasps were tested in the simulator and then a small portion of them were 
validated on a real robot platform equipped with the same hand as the simulator.
In V-REP, for each grasp, the hand pose, the finger joint angles, the object's mesh model, the object's pose, one RGB-D image, and four point clouds taken from different view-points are collected. 

\subsection{Object and data sets}
In Jane Shi's presentation "Automotive General Assembly Part Datasets And their Environment" from General Motors Global R\&D Center, pre-grasp pose datasets were generated independent of the assembly parts. The pre-computed grasp pose datasets represent the robotic hand's ``spatial'' capability that is used to quickly narrow down the search space of searching for a proper grasp when a assembly part is presented. 

Nicolas Alt, Jingyi Xu and Eckehard Steinbach from Technische Universitat Munchen presented a dataset of object models for thin-walled objects in "A dataset of thin-walled deformable objects for manipulation planning." Examples of think-walled objects are bottles, glasses, vases and other containers typically found in households and offices. The models are created using computer models based on a parametric object model generator to resemble real objects. The generator samples in the parameter space to create many variants of an object. Local deformation characteristics and stiffness features of the surfaces can be determined by simulation based on Finite Elements (FEM) using the computer models. The datasets contain the surface and volumetric mesh for each object along with its local stiffness maps. For each object, 50-100 variants with different sizes and stiffness features are created using the parameter generator. The datasets will be available on http://www.lmt.ei.tum.de. 

The presentation "CapriDB - Capture, Print, Innovate: A Low-Cost Pipeline and Database for Reproducible Manipulation Research" is by Florian T. Pokorny, Yasemin Bekiroglu, Karl Pauwels, Judith Bütepage, Clara Scherer, and Danica Kragic from the University of Birmingham, KTH Royal Institute of Technology, and University of California at Berkeley. They presented a novel approach that 3d reconstruct objects using images from a tracked monocular camera. CapriDB - an extensible database were generated using this approach. It currently contains over 40 textured and 3D printable mesh models. The final database was plan to be hosted as a GIT repository. Its current pre-release version of is under development at
http://www.csc.kth.se/~fpokorny/static/capridb/, while the final version will be hosted at http://www.csc.kth.se/capridb/.

\section{Panel Discussion}
The last part of the workshop consisted of a panel discussion. Members of the panel were invited speakers, the workshop organizers but also other researchers with a stake in datasets for manipulation and grasping.

\begin{itemize}
\item Dieter Fox, University of Washington, Seattle, WA, USA
\item Wolfram Burgard, Albert-Ludwigs-University Freiburg, Germany
\item Matt Mason, Carnegie Mellon University, Pittsburgh, PA, USA
\item Abhinav Gupta, Carnegie Mellon University, Pittsburgh, PA, USA
\item Tamim Asfour, Karlsruhe Institute of Technology, Germany
\item Yasemin Bekiroglu, University of Birmingham, UK
\item Yu Sun, University of South Florida, FL, USA
\item Matteo Bianchi, University of Pisa and Istituto Italiano di Tecnologia, Italy
\item Aaron Dollar, Yale University, CT, USA
\item Jeannette Bohg (Moderator), MPI for Intelligent Systems, T{\"u}bingen, Germany
\end{itemize}

There were three main streams of discussion which are summarized in the following sections.

\subsection{Purpose of datasets: Learning vs. Models}
Datasets can be created for different purposes. They can be used to benchmark existing model-based approaches or they can be created to learn models from them. For the latter case, lots of data may be required that can be leveraged by the currently very successful large capacity models.  When used for learning, they also need to be  split into training, test and validation data. This scale of data is difficult to collect on a robotic system. However, we may just not have tried hard enough. Current efforts on collecting large scale datasets on real robotic systems are scarce but existing and not all of them are publicly released  ~\cite{PintoG16,LevinePKQ16,Morales:04,BekirogluKK10,skewness:10}.   
Industry has a competitive advantage over academia as they have the financial resources to collect and label vast amounts of data and also the computational power to learn from it. These datasets are seldom released to the public. This may lead to a shift in what kind of problems are approached in academia which still keep the big picture in mind. 

Manipulation is a very large problem. Even if what we dream that robots are capable of in a few years comes true, it will still be far away from what humans are capable of. There is a huge gulf in front of us even if we see some manipulation skills currently being learned. It is not clear if ‘learning is the answer’.There will be different phases where learning will help us out a lot and new barriers will come up. Research advances by trying a diversity of approaches and being open-minded. There will not be only learning and only modelling, the answer will lie somewhere in-between  Interesting questions include how can learning and models be combined to learn from much less data? How to re-use learned models from simpler problems to more complex chains of problems?
 
\subsection{How to benchmark with hardware?}
Deep learning techniques have not impacted Robotics as much as for example Computer Vision and Natural Language Processing. This may be because the structure of the data is very different: it results from a sequential decision making process in which the next datapoint is influenced by the previous movement command. This is different from the current structure of the majority of datasets which contains identically and independently distributed (i.i.d.) data points. Some problems in robotics can be cast such that learning from such a dataset if possible (e.g. for grasping~\cite{PintoG16,Popovic:11,KapplerBS15,Goldfeder:2009,DeepGrasping:2014,mahlerdex}). However, many other problems require testing with actual hardware. This raises the question how more research labs can have access to this expensive equipment to ‘democratise' Robotics research. Can researchers have access to platforms to run their work and test it? Can we get hundreds of robots bought by the community and in a central area? It is still much more convenient to have a setup in your own lab. Simulation may bridge this gap. All this would require the definition of a standardised test environment as for example existing at KIT. Yasemin Berikoglu suggests to have two robots where one places the objects in a repeatable way and another robots grasps the objects using different approaches which are benchmarked  Other scenarios include elderly care, warehouse, and industrial automation. It would however be useful to define the most challenging scenarios by a group of experts. 

Having a standardised test environment makes it easier to compare approaches and also to share modules. The best development within Computer Vision is that code is always released to make experiments reproducible. Although some approaches may score best on benchmarks, in practice approaches that scored second or third are used much more because they actually work. 
In standardised test scenarios, there is however the danger of overfitting to scenarios. This is probably not avoidable but can be mitigated by making problems harder and harder. It is also important to make the training and test datasets or scenarios as separate as possible to really test generalisation. Also these stages of difficulty and the test scenarios can be defined best by a group of experts. Even with some overfitting, there will still be some fundamental insights to be gained. And given the complexity of the problem and the current performance of robotics systems, making the scenario or test sets more difficult is very easy. 

Another very difficult question is what should be benchmarked once a test scenario is defined. Should modules be tested or systems? While modules may be easier to compare on datasets (perception, learning) that capture some subproblems, it is unclear how they perform when integrated into a whole system. Especially in robotics, experience has shown that the success of a whole system does not critically depend on one modules but on how different modules are connected together. The more interesting/diverse the test, the more difficult it is to measure progress and success. How can we attribute progress and success for a specific tasks to elements of systems? Making progress in this area if very important for progressing in robotics which is a science that revolves around building systems. The metrics to measure progress could for example include speed of execution and speed of learning. In general, it would be good to report a mixture of quantitative and qualitative results. Important is also the report of failures. Other things missing is a coherent baseline that may consist of classic modules and is put together in a classic sense-plan-act architecture.

\subsection{Collection of datasets}
There is also currently a lack of coherence in the collection of datasets. One effort to make this process more structured is done by Matteo Bianchi and Minas Liarokapis, who collect datasets on human and robot hands in one place (HandCorpus: www.handcorpus.org), an open access repository freely accessible (no login or password required). All these have different formats and collect different attributes and different experimental paradigms. Currently, this is unified by hand.

\section{Conclusion}
In general the workshop was very successful. The half day workshop attracted over one hundred attendances. The nine invited talks and twelve interactive presentations represent the focuses and trends of latest datasets. Many datasets were collected for evaluation benchmarks or data-driven learning (machine learning or deep learning). Several datasets look into previously under-explored modalities such as tactile signal, friction, or interactive force and torque. Many approaches have been developed to carry out automatic data collections to improve efficiency and lower the cost. However, so far, obtaining a large dataset of human/robot grasping and manipulation motions with a large number of objects and rich modalities is still very challenging. High quality, coherent and big datasets will be vital to learning scale and speed in the robotics community. 

\section*{Acknowledgement}
The workshop was supported by the IEEE Robotics and Automation Society (RAS) Technical Committee (TC) on Robotic Hands, Grasping and Manipulation, IEEE RAS
TC on Mobile Manipulation, IEEE RAS TC on Computer \& Robot Vision, IEEE RAS TC on Human Movement Understanding, and IEEE RAS TC on Haptics.

\bibliographystyle{plain}
\bibliography{2016_WS_doc}

\end{document}